\documentclass[letterpaper,10pt,conference]{ieeeconf}

\IEEEoverridecommandlockouts                              

\usepackage{graphicx}
\usepackage{amsmath,amssymb} 
\usepackage{color}
\usepackage{multicol}
\usepackage{bm}
\usepackage{url}
\usepackage{multirow}
\usepackage{times}
\usepackage{gensymb}
\usepackage{cite}
\usepackage{times}
\usepackage{epsfig}
\let\labelindent\relax
\usepackage{enumitem}
\usepackage{mathrsfs}
\usepackage{booktabs}
\usepackage{mathtools, cases}
\usepackage{microtype}
\usepackage[pagebackref=true,breaklinks=true,letterpaper=true,colorlinks,bookmarks=false]{hyperref}

\title{\LARGE \bf A*3D Dataset: Towards Autonomous Driving in Challenging Environments} %

\usepackage[font=footnotesize,labelformat=simple]{subcaption}

\newcommand{\ie}{\textit{i}.\textit{e}.}

\author{
  Quang-Hieu Pham$^\star$$^{1}$, Pierre Sevestre$^\star$$^{2}$, Ramanpreet Singh Pahwa$^{3}$, Huijing Zhan$^{3}$, Chun Ho Pang$^{3}$, \\ Yuda Chen$^{3}$, Armin Mustafa$^{4}$, Vijay Chandrasekhar$^{3}$, and Jie Lin$^\dagger$$^{3}$ %
  \thanks{$^{1}$Quang-Hieu Pham is with Singapore University of Technology and Design, Singapore {\tt\small quanghieu.pham@mymail.sutd.edu.sg}}%
  \thanks{$^{2}$ Pierre Sevestre is with CentraleSupélec, France {\tt\small pierre.sevestre@student.ecp.fr}}%
  \thanks{$^{3}$ Ramanpreet S. Pahwa, Huijing Zhan, Chun Ho Pang, Yuda Chen, Vijay Chandrasekhar. and Jie Lin are with the Institute for Infocomm Research (I$^2$R), A$^*$STAR, Singapore {\tt\small \{ramanpreet$\_$pahwa, zhan$\_$huijing, vijay, pang$\_$chun$\_$ho, chen$\_$yuda, lin$-$j\}@i2r.a-star.edu.sg}}
  \thanks{$^{4}$ Armin Mustafa is with CVSSP, University of Surrey, UK {\tt\small a.mustafa@surrey.ac.uk}}
  \thanks{$^\dagger$This work is supported by the Agency for Science, Technology and Research (A$^*$STAR) under its AME Programmatic Funds (Project No.A1892b0026). Corresponding author: Jie Lin.}
  \thanks{$^\star$This work is done when Pierre and Hieu interned at I$^2$R, A$^*$STAR. Both authors contributed equally.}%
}

\graphicspath{{./images/}}

\begin{document}

\maketitle
\pagestyle{empty}

\begin{abstract}
  With the increasing global popularity of self-driving cars, there is an immediate need for challenging real-world datasets for benchmarking and training various computer vision tasks such as 3D object detection. 
  Existing datasets either represent simple scenarios or provide only day-time data.
  In this paper, we introduce a new challenging A*3D dataset which consists of RGB images and LiDAR data with significant diversity of scene, time, and weather.
  The dataset consists of high-density images ($\approx~10$ times more than the pioneering KITTI dataset), heavy occlusions, a large number of night-time frames ($\approx~3$ times the nuScenes dataset), addressing the gaps in the existing datasets to push the boundaries of tasks in autonomous driving research to more challenging highly diverse environments. 
  The dataset contains $39\text{K}$ frames, $7$ classes, and $230\text{K}$ 3D object annotations. An extensive 3D object detection benchmark evaluation on the A*3D dataset for various attributes such as high density, day-time/night-time, gives interesting insights into the advantages and limitations of training and testing 3D object detection in real-world setting.
\end{abstract}

\section{Introduction}
Self-driving cars largely exploit supervised machine learning algorithms to perform tasks such as 2D/3D object detection~\cite{shi2019pointrcnn,lang2019pointpillars}.
Existing autonomous driving datasets represent relatively simple scenes with low scene diversity, low density and are usually captured in day-time~\cite{geiger2013kitti,patil2019h3d}.
Training and benchmarking detection algorithms on these datasets lead to lower accuracy when applying them in real-world settings, making it unsafe for the public.
With the increasing use of autonomous vehicles, it is important for these datasets to not only represent different weather and lighting conditions but also contain challenging high-density diverse scenarios.

\begin{figure}[t!]
  \centering
  \includegraphics[width=\linewidth]{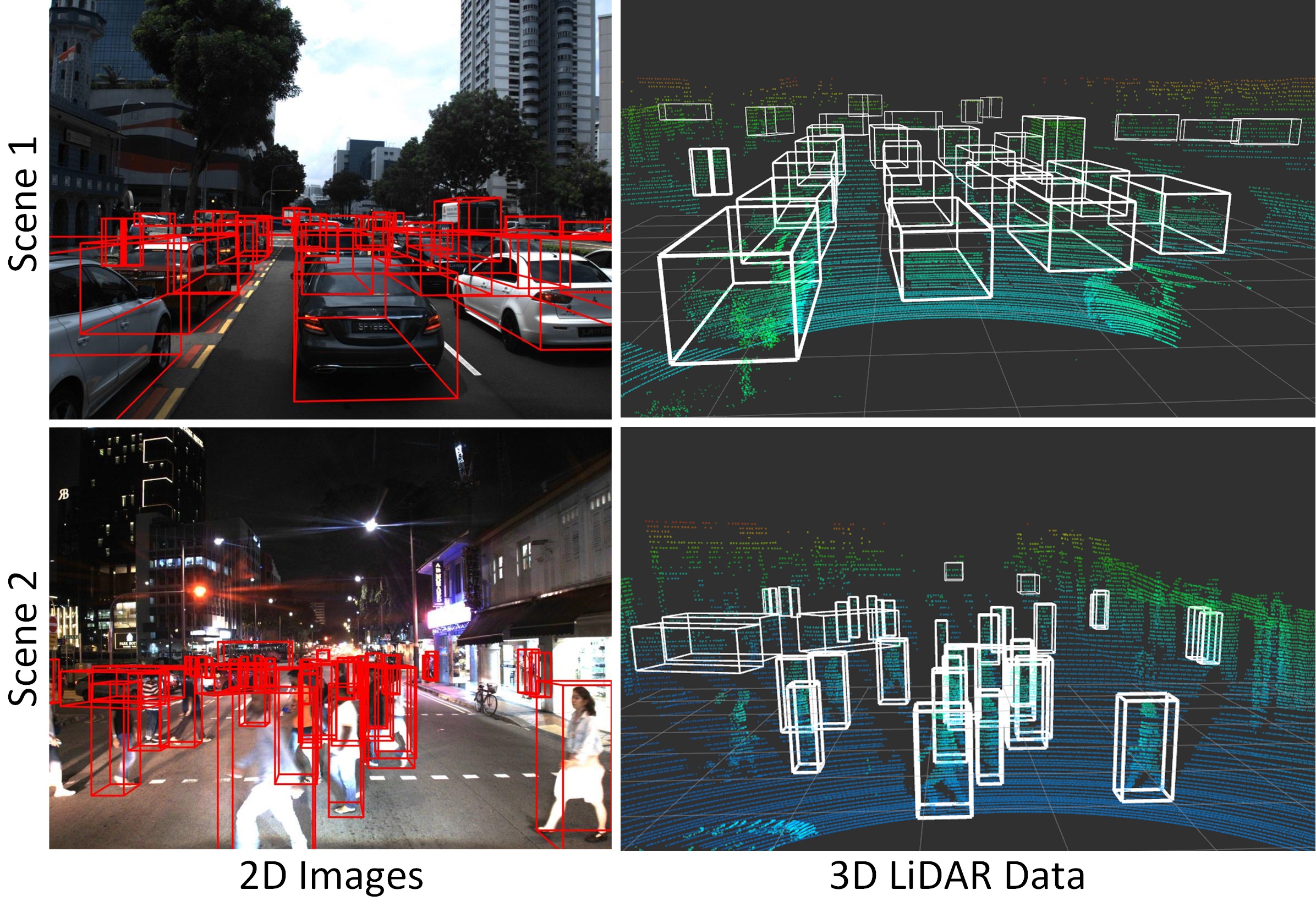}
  \caption{
    Samples from the proposed A*3D dataset with RGB images and their corresponding LiDAR data. The two scenes captured in the evening and at night demonstrate high object-density in the environment.
  }
  \vspace{-0.2in}
  \label{fig:motivation}
\end{figure}

\begin{table*}[t]
  \centering
  \begin{tabular}{lccccccccc}
    \toprule
    \multirow{2}*{Dataset} & \# Annotated & Night & \# 3D & \# 3D boxes & Density & Annotated & Driving  & \multirow{2}*{Location} & Scene \\
    & LiDAR & ratio ($\%$) & boxes & (front view) & (front view) & freq. (Hz) & speed (km/h) & & diversity \\
    \midrule
    KITTI'12~\cite{geiger2013kitti} & 15k & 0 & 80k & 80k & 5.3 & - & - & Karlsruhe & Low \\
    KAIST'18~\cite{choi2018kaist} & 8.9k & - & 0 & 0 & 0 & - & - & Seoul & Low \\
    H3D'19~\cite{patil2019h3d} & 27k & 0 & 1M & - & - & 2 & - & San Francisco & Low \\
    nuScenes'19~\cite{caesar2019nuscenes} & 40k & 11.6 & \emph{1.2M} & \textbf{\emph{330k}} & \textbf{\emph{9.7}} & 2 & 16 & Boston, Singapore & Low \\
    A*3D'19 & 39k & \textbf{30} & 230k & 230k & 5.9 & \textbf{0.2} & \textbf{40-70} & Singapore & \textbf{High} \\
    \bottomrule
  \end{tabular}
  \caption{
    Comparison of A*3D with existing datasets --- $2^{nd}$ column represents the number of annotated LiDAR frames. Each LiDAR frame corresponds to one or more camera images.
    $3^{rd}$ column is the proportion of frames recorded during night time,
    $4^{th}$ column is the number of 3D bounding boxes captured by frontal camera only,
    $5^{th}$ column is the average number of objects per frame in front view,
    $6^{th}$ column is the number of frames annotated per second,
    and $7^{th}$ column summarizes the diversity of recorded scenes (spatial coverage, weather, timing, road type, surroundings, etc.).
    Numbers in fourth row for nuScenes highlighted in italics are averaged over $34,149$ trainval frames as released by nuScenes organizers.
  }
  \label{tab:dataset}
\vspace*{-5mm}
\end{table*}

To address the issues, we introduce a new diverse multi-modal dataset, called \emph{A*3D}, for autonomous driving in challenging environments.
An illustration of our dataset is shown in Figure~\ref{fig:motivation}.
A*3D has $230\text{K}$ human-labeled 3D object annotations in $39,179$ LiDAR point cloud frames and corresponding frontal-facing RGB images captured at different times (day, night) and weathers (sun, cloud, rain).
The spatial coverage of A*3D nearly spans the whole Singapore country, with diverse scenes such as highways, neighborhood roads, tunnels, urban, suburban, industrial, car parks, etc.
A*3D dataset is a step forward to make autonomous driving safer for pedestrians and the public in the real world.

An overall comparison of existing autonomous driving datasets with LiDAR point clouds is shown in Table~\ref{tab:dataset}.
It is worth noting that KITTI~\cite{geiger2013kitti}, the pioneering multi-modal dataset, consists of only day-time frames and on average $5.3$ object instances per frame.
The nuScenes dataset~\cite{caesar2019nuscenes} consists of $9.7$ object instances per frame (object density is computed with the frontal view only for a fair comparison) and $11.6\%$ night-time frames.
Despite of the higher object-density, the diversity of the nuScenes data is limited by its small spatial coverage (a small region in Singapore and Boston Seaport, USA), low driving speed ($16$ km/h), and high annotating frequency ($2$ Hz).
Compared to these existing datasets, A*3D is more challenging as it comprises of $30\%$ night-time frames, $5.9$ object instances per frame with a large portion of the annotated objects, mainly vehicles, being heavily occluded and exhibits much higher scene diversity.

Deploying autonomous vehicles in real environments requires the detection of objects such as vehicles, pedestrians, and cyclists in 2D/3D on RGB images and LiDAR point clouds.
Numerous 3D object detection methods~\cite{shi2019pointrcnn,ku2018avod,lang2019pointpillars} have been proposed on existing autonomous driving datasets KITTI~\cite{geiger2013kitti}, nuScenes~\cite{caesar2019nuscenes} and H3D~\cite{patil2019h3d}.
In this paper, we utilize A*3D dataset to perform comprehensive 3D object detection benchmarking of the state-of-the-art algorithms under different configurations in varied challenging settings including night-time, high object density and heavily occluded objects.
The contributions of the paper are:
\begin{itemize}
\item A new challenging A*3D dataset is proposed for autonomous driving in the real world with highly diverse scenes, attributed to the large spatial coverage of the recording, high driving speed, and low annotation frequency.
\item A*3D dataset contains $17\%$ frames with high object-density ($>10$ object instances), $25\%$ frames with heavy occlusion, and $30\%$ frames captured during night-time. This enables the community to study 3D object detection in more challenging environments, supplementing current state-of-the-art autonomous driving datasets.
\item An extensive 3D object detection benchmark performance evaluation is demonstrated on the A*3D dataset under different configurations designed for evaluating the effects of density and lighting conditions. Insights drawn from our performance analysis suggest interesting and open questions for future research.
\end{itemize}

The A*3D dataset and benchmark evaluation are released at \url{https://github.com/I2RDL2/ASTAR-3D} for non-commercial academic research purposes only.

\section{Related Datasets}\label{sec:related_work}
Datasets play a key role in Autonomous Driving research. However, most of these datasets provide data for driving in day-time and represent simple scenes with low diversity~\cite{geiger2013kitti,patil2019h3d}.
A majority of these datasets focus on 2D RGB images and annotations of street data such as Cityscapes~\cite{cordts2016cityscapes}, Camvid~\cite{brostow2008camvid}, Apolloscape~\cite{huang2018apolloscape}, Mapillary~\cite{neuhold2017mapillary} and BDD100K~\cite{yu2018bdd100k}.
Some datasets were introduced that focus only on pedestrian detection~\cite{enzweiler2008monocular,ess2008mobile,dollar2011pedestrian}.

In recent years, various multi-modal datasets with RGB images and 3D LiDAR data were released.
The first key multi-modal dataset that greatly advanced 3D object detection is KITTI, which consists of point clouds from LiDAR and stereo images from front-facing cameras with small-scale annotations, i.e. $80K$ 3D boxes for $15K$ frames.

Another multi-modal dataset, H3D~\cite{patil2019h3d}, contains $27K$ frames with $1M$ 3D boxes in the full $360\degree$ view. Despite the higher number of annotations compared to KITTI, H3D manually annotated one-fifth of the frames and propagated the human-labeled annotations to the remaining frames using a linear interpolation technique, which leads to reduced annotation quality and accuracy. Two recently relased datasets, KAIST~\cite{choi2018kaist} and nuScenes~\cite{caesar2019nuscenes}, include data captured in the night time and rainy weather. While nuScenes provides $1.2$M 3D object annotations in $360\degree$ view, KAIST annotations are in 2D images only.
One thing to note here is that the number of annotations is usually much higher in $360\degree$, such as H3D and nuScenes, compared to frontal-view datasets, such as KITTI and the proposed A*3D dataset. However, when only considering the frontal annotations, these numbers drop considerably, for example, the number of 3D boxes is reduced from $1.2M$ to $330k$ for nuScenes (Table~\ref{tab:dataset}).

A*3D dataset is a frontal-view dataset which consists of both day and night-time data with 3D annotations, unlike KITTI, H3D (only day-time data), and KAIST (only 2D).
There are major differences in driving and annotation planning between A*3D and nuScenes datasets, as shown in Table~\ref{tab:dataset}.
First, the nuScenes data is collected from a very small part of Singapore (One-North, Queenstown and Holland Village) and Boston Seaport, USA, while the driving routes for A*3D cover almost the entire area of Singapore (Fig.~\ref{fig::map_AV}).
Hence, our driving scenarios are much more comprehensive.
Second, the driving speed for A*3D is much higher than nuScenes (40-70 km/h vs. 16 km/h), leading to discrepancies in the captured data because highly dynamic environment makes it more challenging for detection tasks.
Third, the annotating frequency for A*3D is 10 times lower than nuScenes (0.2Hz vs. 2Hz), which increases the annotation diversity.
Hence A*3D dataset is highly complex and diverse compared to nuScenes and other existing datasets, shaping the future of autonomous driving research in challenging environments.

\begin{section}{The A*3D Dataset}

  \begin{subsection}{Sensor Setup}
    We collect raw sensor data using the A*STAR autonomous vehicle, as shown in Fig.~\ref{fig::i2rav}. This vehicle is equipped with the following sensors:
\begin{itemize}
    \item Two PointGrey Chameleon3 USB3 Global shutter color
      cameras (CM3-U3-31S4C-CS) with $55$Hz frame rate, $2048 \times 1536$ resolution, Sony IMX265 CMOS sensor and $57.3\degree$ field-of-view (FoV).
    \item A Velodyne HDL-64ES3 3D-LiDAR with 10Hz spin-rate, 64 laser beams, $0.17\degree$ horizontal and $0.33\degree-0.5\degree$  vertical angular resolution. The LiDAR is capable of $\le 2$ cm distance accuracy and measures 133K points/rev with vertical and horizontal FoV of $360\degree$ and $26.8\degree$ respectively and a reliable range of $0.9m$-$100m$
\end{itemize}
Sensor data is recorded using Ubuntu $16.04$ LTS machine with a quad-core Intel $i7-6700K$ $2.4$ GHz Quad-Core processors, $16GB$ DDR4L memory, and a RAID $0$ array of a $1$TB internal SSD and a $2$TB external Samsung SSD via USB3. The cameras were connected using two USB3 ports on dedicated PCI-E USB card and the Velodyne LiDAR was connected using the LAN port. We used ROS Kinetic Kame to save the raw data as bagfiles in one minute intervals.
\begin{figure}[t!]
  \centering
      \includegraphics[width=\linewidth]{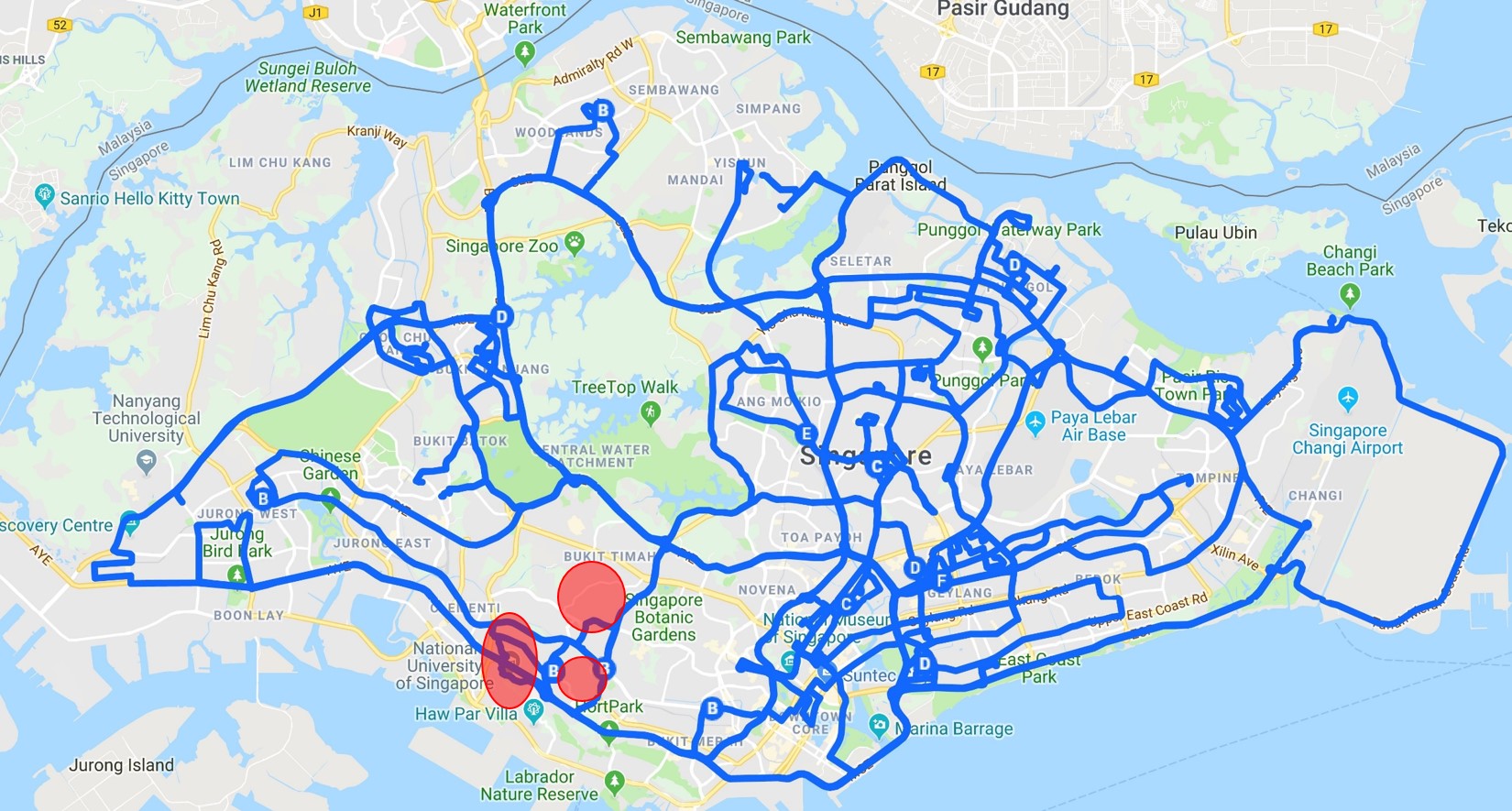}
      \caption{The driving routes and the spatial coverage of the A*3D dataset overlaid on Google map. Our dataset encompasses the entire Singapore while nuScenes\cite{caesar2019nuscenes} only covers a small portion of Singapore roads (highlighted in red).}
  \label{fig::map_AV}
\end{figure}    
  \end{subsection}
  \begin{subsection}{Sensor Synchronization and Calibration}
    All the sensor are synchronized using Robot Operating System (ROS), and carefully calibrated by using a mobile checkerboard structure and a simultaneous multi-modal calibration method between the color Chameleon3 cameras (camera-to-camera) and the color-Velodyne devices (camera-to-LiDAR).

We follow KITTI's \cite{Geiger2012} frame of reference by expressing extrinsic coordinates in LiDAR's point of view with: x = right, y = down, z = forward as our coordinate system.
We hire a third party company to measure the intrinsic LiDAR calibration parameters used for the entire data collection.
We use a standard checkerboard that is moved around the scene to collect sample images for camera intrinsic and extrinsic calibration along with camera-LiDAR calibration.

Particularly, the OpenCV \cite{opencv_library} standard calibration toolbox is used to estimate the intrinsic and extrinsic camera parameters. A similar framework can be used for camera-LiDAR calibration provided that we have accurate 3D corner measurements in the LiDAR point cloud data. Camera-LiDAR cross-calibration results obtained from Geiger \cite{Geiger2012} and CMU \cite{unnikrishnan2005fast} toolboxes gave a high mean reprojection error of $>5$ pixels.
To resolve this issue, inspired from \cite{WANG2017Lidar_camera_cali}, a checkerboard was used for calibration, such that the there are no whitespaces around the checkerboard. This makes it easier to locate the four 3D corners of the checkerboard in the LiDAR dataset and extract internal 3D corners of the checkerboard pattern in the point cloud. This approach resulted in a reprojection error of $<2$ pixels that is much lower than above described toolboxes. 
  \end{subsection}
  \begin{subsection}{Data Collection}
\begin{figure}[t!]
\centering
\includegraphics[width=0.33\textwidth, height=0.40\textwidth]{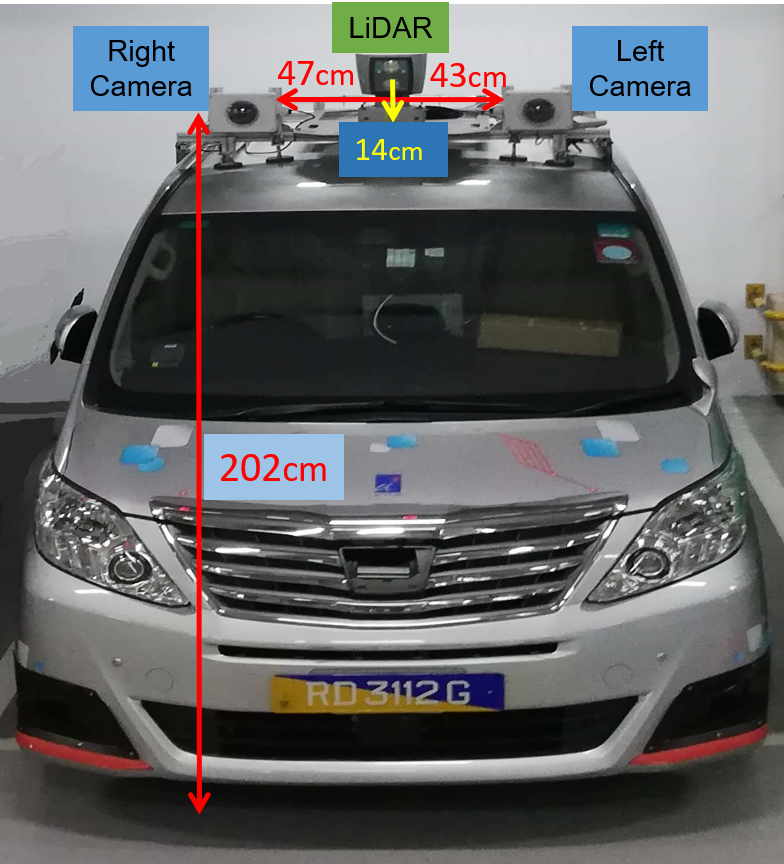}
\caption{Sensor setup for A*3D data collection vehicle platform. The A*STAR autonomous driving vehicle consists of a rotating Velodyne LiDAR and two color PointGrey Chameleon3 cameras placed on either side of the LiDAR.}
\label{fig::i2rav}
\end{figure}
As highlighted in Fig.~\ref{fig::map_AV}, the driving routes are overlaid on Google map and the data collection covers entire Singapore including highways, neighborhood roads, tunnels, urban, suburban, industrial, HDB car parks, coastline, etc. Driving generally started in the afternoon and ended before midnight, in a wet season month, March 2018, and a dry season month, July 2018. We collected around 4-hours worth of driving data per day on an average. To reflect real-world driving scenarios, the driving speed ranges from $40$ km/h to $70$ km/h, following the road regulations in Singapore. As mentioned in Sec.~\ref{sec:related_work}, this enables us to construct a highly challenging autonomous driving dataset with increased scene diversity in time, weather and data quality.

  \end{subsection}
  \begin{subsection}{Data Annotation}
    Unlike the high annotation frequency ($2Hz$) used by the nuScenes and H3D, we uniformly select $1$ LiDAR frame per $5$ seconds (i.e. annotating frequency at $0.2Hz$) from $55$-hours driving data, resulting in $39,179$ frames. For each LiDAR frame, the nearest neighbor camera frame is identified based on their ROS timestamps. Our domain expert data annotation partner marked 3D bounding boxes directly on the LiDAR point cloud for $7$ object classes (Car, Pedestrian, Van, Cyclist, Bus, Truck, and Motorcyclist), as long as the objects appear in the corresponding camera FoV. Each bounding box is associated with $2$ attributes - occlusion and truncation. Occlusion is categorised into $4$-levels (fully visible, occluded less than $50\%$, occluded more than $50\%$, unknown), and truncation contains $3$-levels (not truncated, truncated less than $50\%$, truncated more than $50\%$).

    With the accurate camera-LiDAR calibration, 3D bounding box in the point cloud is projected to the camera image plane and visualized as 2D and 3D bounding box, to verify that box localization and size are precise on both point clouds and camera images. As far away or heavily occluded objects are easy to miss, our partner paid particular attention to frames with dense object instances, ensuring there are no missing object annotations in such scenarios.
    \end{subsection}
  \begin{subsection}{Statistics}
    
\begin{figure}[t!]
    \centering
    \includegraphics[width=\linewidth]{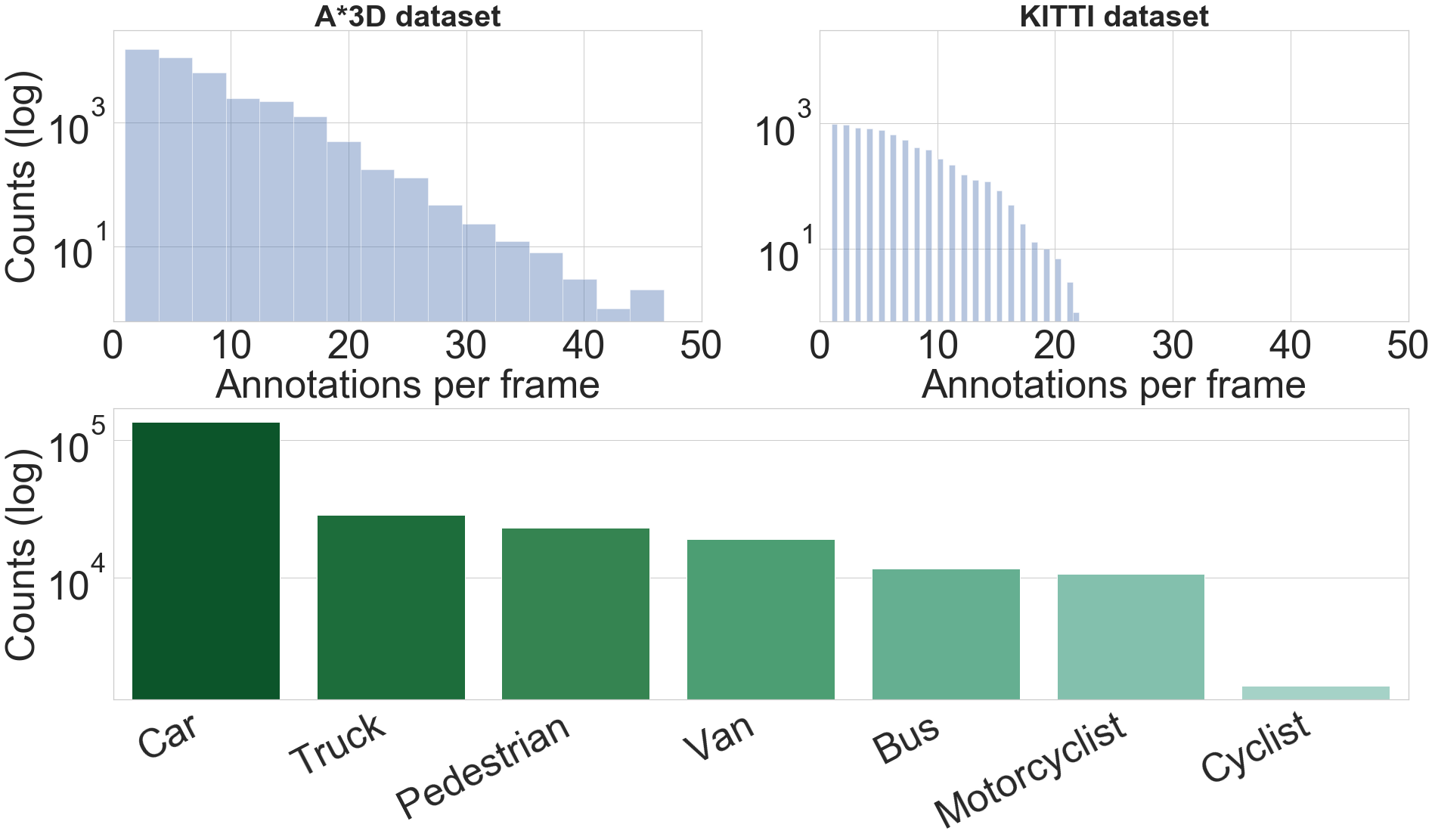}
    \caption{Top: Comparison of the number of object per frame for the KITTI dataset (right), and A*3D dataset (left). Bottom: Number of annotation per category for A*3D.}
    \label{fig:class_point_per_frame}
\end{figure}

\begin{figure}[t!]
    \centering
    \includegraphics[width=\linewidth]{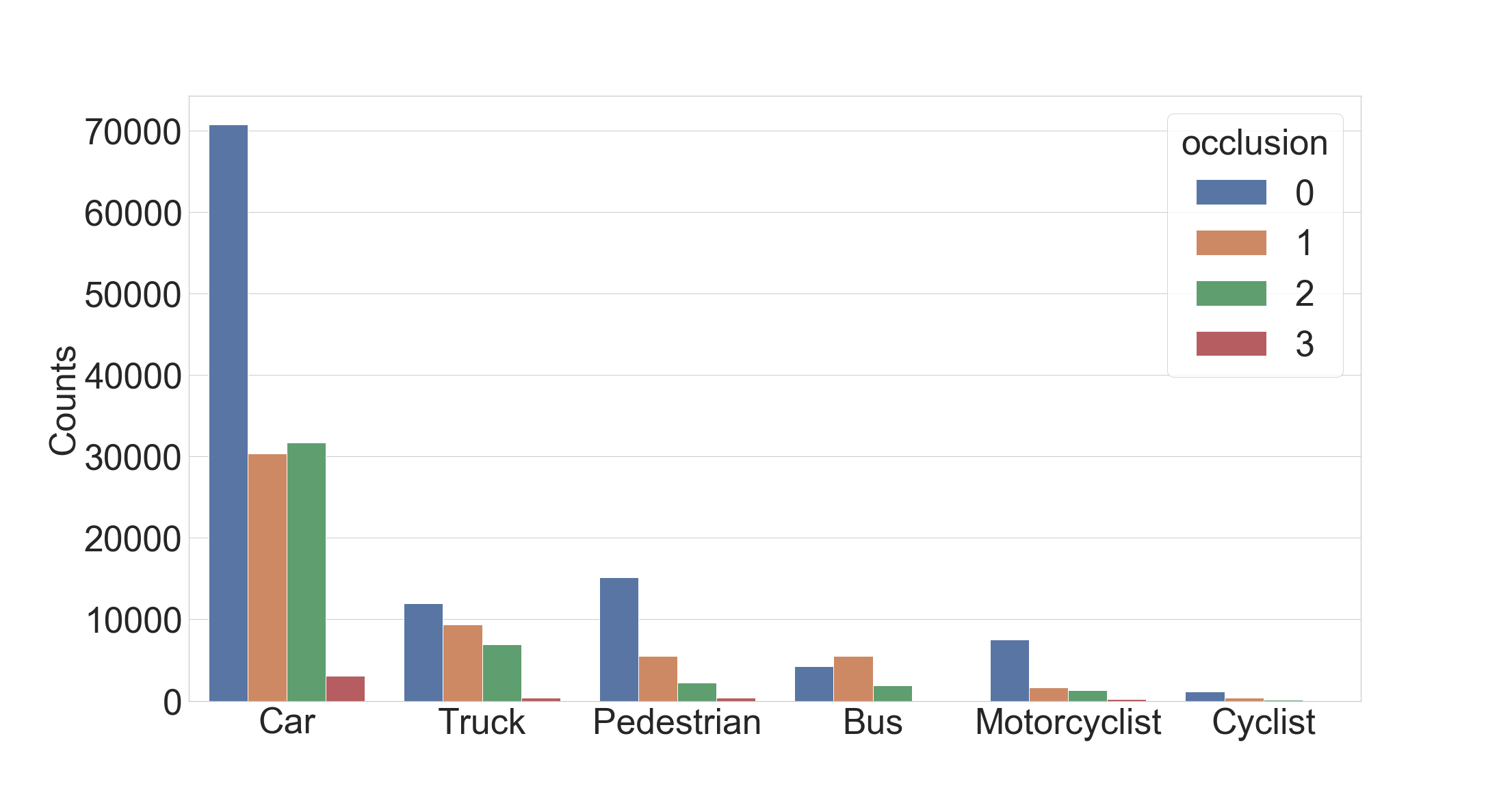}
    \caption{Occlusion level per class. 0: fully visible, 1: occluded less than 50\%, 2: occluded more than 50\%, 3: unknown.}
    \label{fig:occlusion_plot}
\end{figure}

\begin{figure}[t!]
    \centering
    \includegraphics[width=\linewidth]{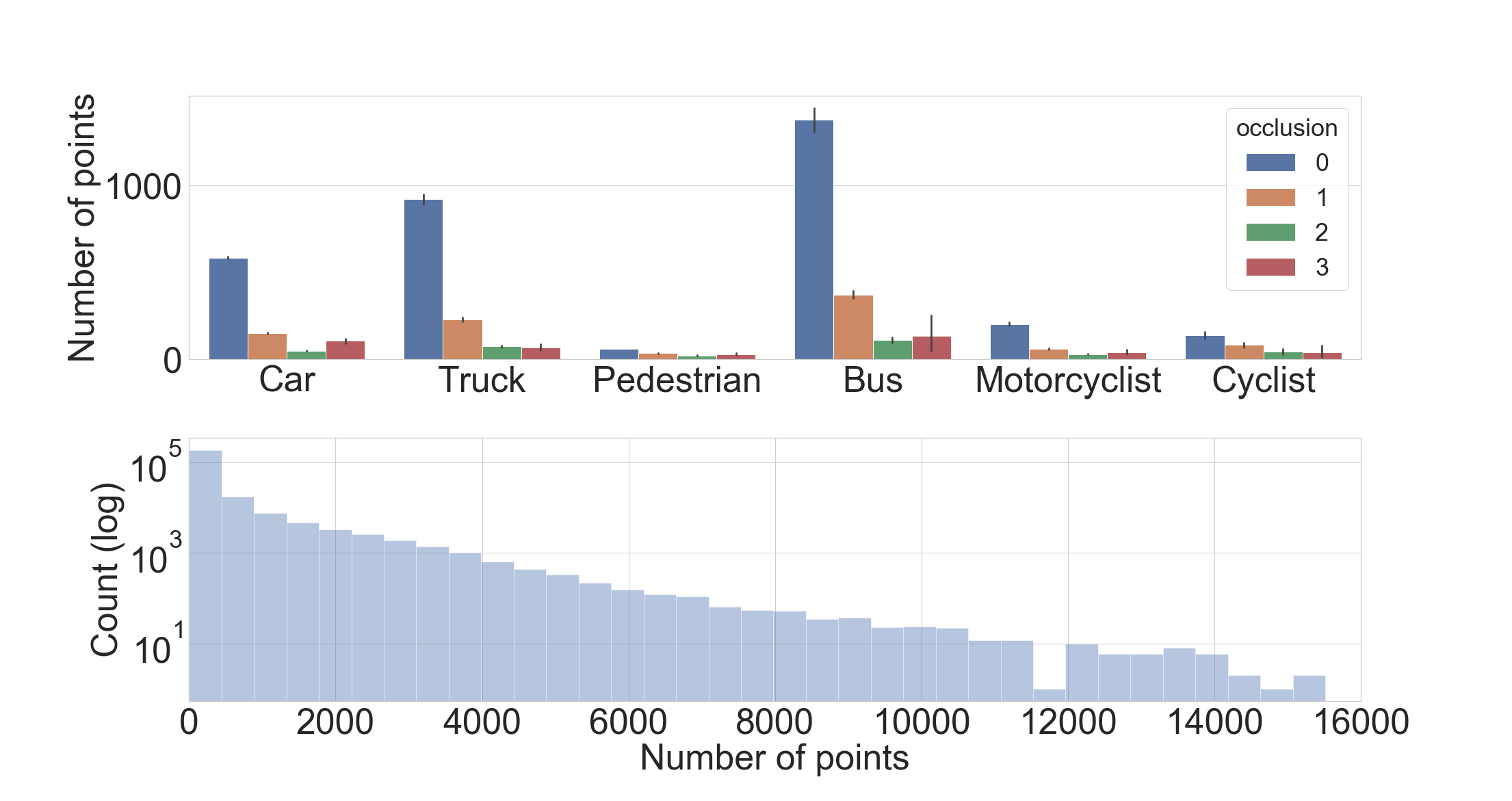}
    \caption{Top: Average number of points inside the bounding box of each class, as a function of occlusion. Bottom: Log number of points within bounding box.}
    \label{fig:occlusion_point_log_points}
\end{figure}

\begin{figure}[t!]
    \centering
    \includegraphics[width=\linewidth]{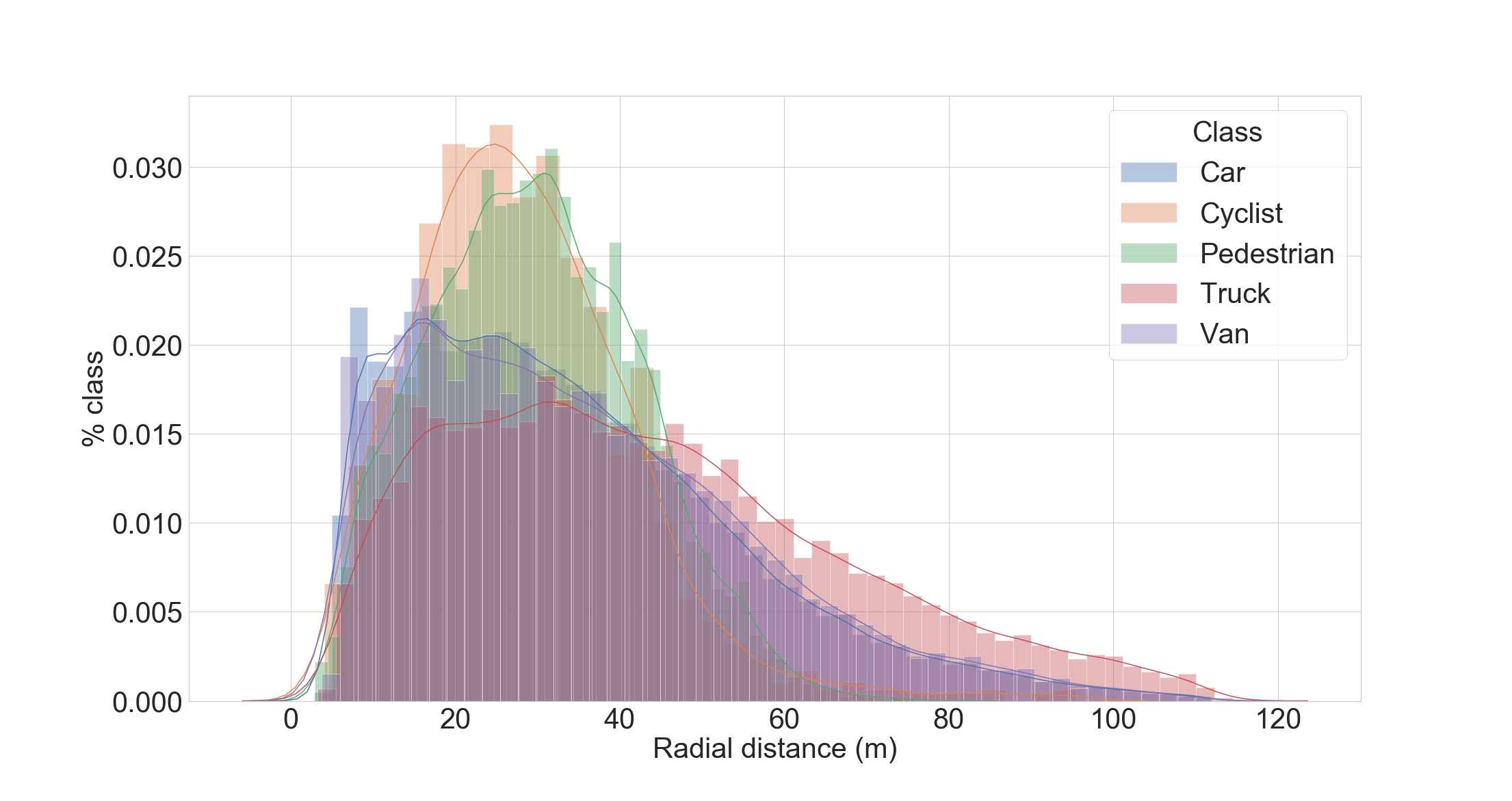}
    \caption{Radial distance in meter from the center of box to the LiDAR sensor.}
    \label{fig:rdist_to_lidar}
\end{figure}

\begin{figure}[t!]
    \centering
    \includegraphics[width=\linewidth]{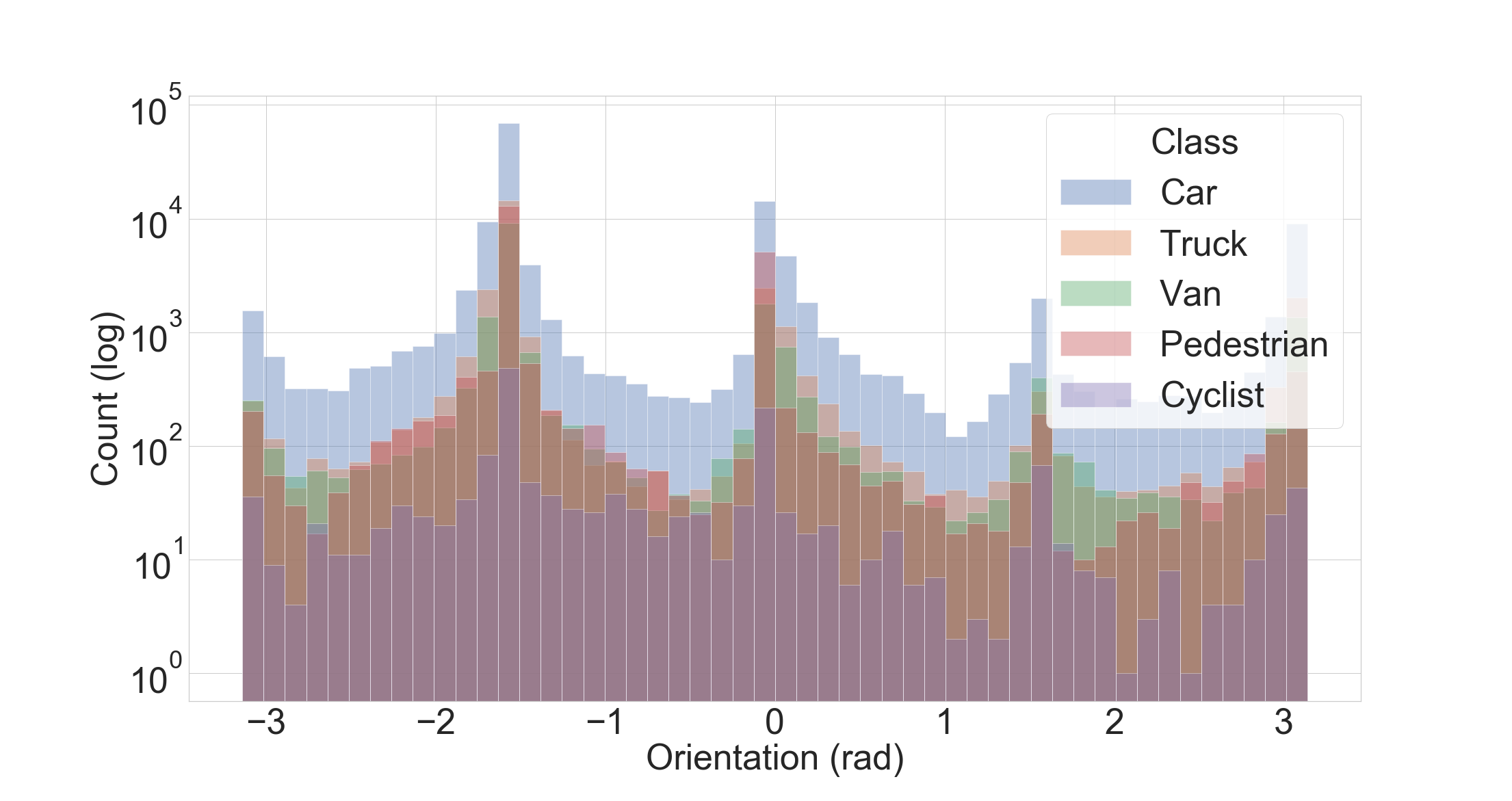}
    \caption{Distribution of object orientation per class.}
    \label{fig:orientation_I2RAV}
\end{figure}

\begin{figure}[t!]
    \centering
    \includegraphics[width=\linewidth]{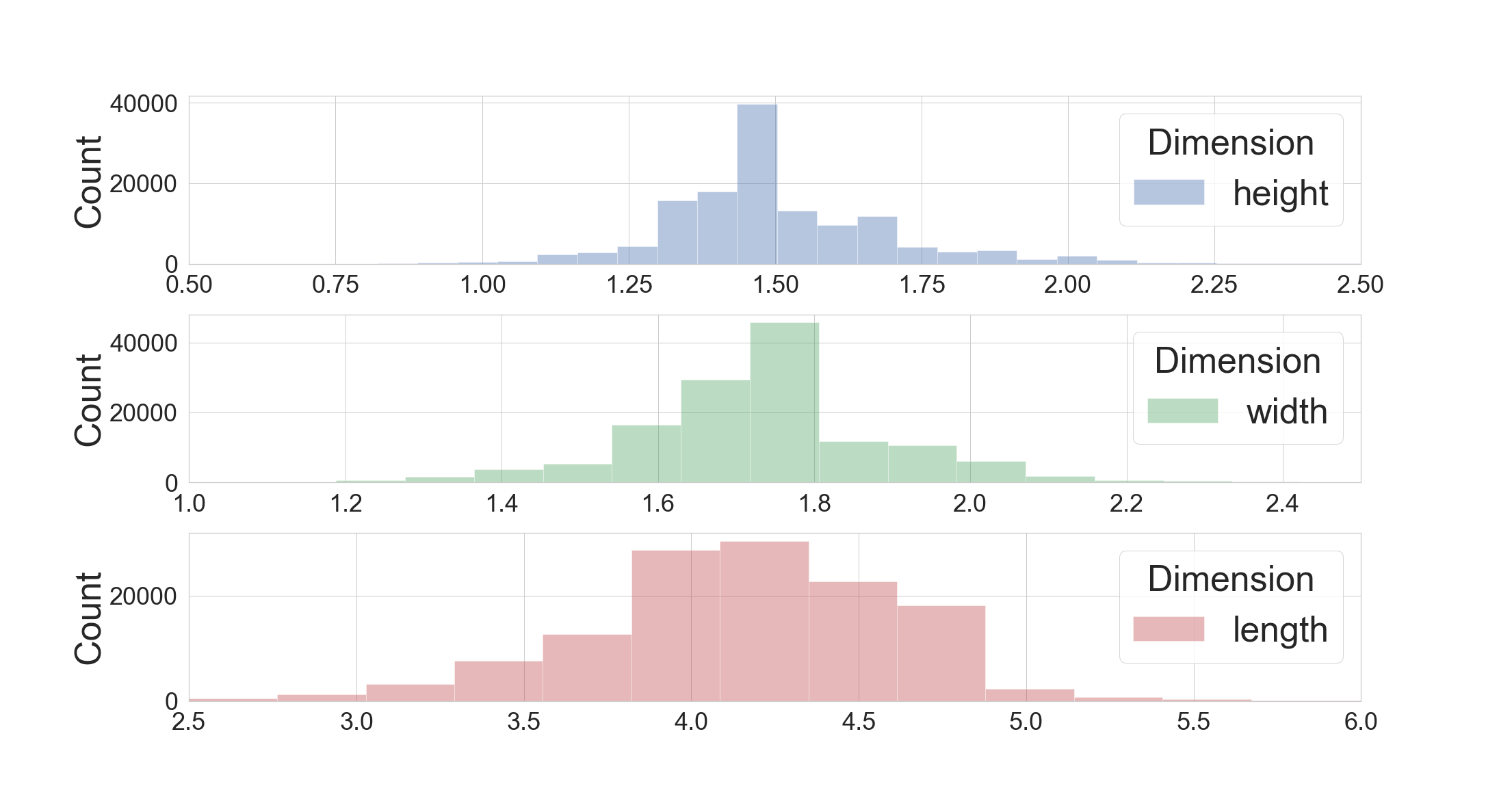}
    \caption{Box dimensions (height [m], width [m], length [m]) for the Car class.}
    \vspace{-5mm}
    \label{fig:car_hwl}
\end{figure}

The A*3D dataset comprises $7$ annotated classes corresponding to the most common objects in road scenes - vehicles (Car, Van, Bus, Truck), pedestrians, cyclists, and motorcyclists, as shown in Fig.~\ref{fig:class_point_per_frame} (Bottom).We provide brief insights into our dataset in this section. 

The proposed dataset contains high-density frames unlike KITTI, as evident from Fig. \ref{fig:class_point_per_frame} (Top). The number of annotations per frame for A*3D dataset are much higher than the KITTI dataset.
Moreover, A*3D dataset consists of heavily occluded frames as shown in  Fig.~\ref{fig:occlusion_plot}, where about half of the vehicles are partially or highly occluded.
Both occlusion and distance significantly impact the number of points contained in objects impacting detection accuracy in a given scene. Highly occluded objects are difficult to detect as the number of points lying inside the bounding boxes are drastically reduced. While fully visible objects contain more than $500$ points on an average, heavily occluded objects contain $40$ points only (Fig. \ref{fig:occlusion_point_log_points} (Top)). Moreover, objects $20$m away from LiDAR usually consist of more than $600$ points. This number drops to less than $40$ points for objects more than $50$m away from LiDAR. 
The average radial distance to the camera is $34m$ for the car class, which is consistent over different classes (Fig. \ref{fig:rdist_to_lidar}).
The wide range of LiDAR sensor allows capturing pedestrians and vehicles up to $60m$ and $100m$ respectively.
Recording in a city environment leads to imbalanced orientation distribution, as shown in Fig. \ref{fig:orientation_I2RAV}, with $60\%$ of the objects lying within [-10\degree, +10\degree] of LiDAR's FoV.
Fig.~\ref{fig:car_hwl} presents a reasonable size distribution (height, width, and length) of the most popular Car objects.
Recorded over multiple days, the dataset contains diverse weather and time conditions, including $30\%$ of night-time frames.

  \end{subsection}
\end{section}

\section{Benchmarking}
In this section, we benchmark representative 3D object detection methods on the A*3D dataset. More specifically, we design different train/validation data configurations on A*3D, in order to investigate how detection accuracy is impacted in challenging environments with high object-density and night-time data. Our results indicate several limitations of existing 3D object detectors in a challenging environment and suggest possible directions for future research.

We define a standard validation set for evaluating all of the experiments in the following section.
This validation set should have various desirable properties for a challenging benchmark, such as highly occluded scenes, day/night frames.
In more details, we carefully select $1,500$ LiDAR frames from the entire A*3D dataset as the validation set.
All frames with high object-density ($\textgreater{}10$) as well as heavy occlusion ($\textgreater{}1$).
On the other hand, the validation set consists of $1,000$ day-time samples and $500$ night-time samples keeping the same ratio of day-night samples as in the main dataset.
Following the protocols defined in KITTI~\cite{geiger2013kitti}, we decompose the detection difficulty on the evaluation set into 3 levels, \ie{} \emph{Easy}, \emph{Moderate}, \emph{Hard}, which is determined by bounding box size, occlusion, and truncation.
We describe the design of training data in the respective subsections.

We select widely used state-of-the-art 3D object detectors that are publicly available online, namely PointRCNN~\cite{shi2019pointrcnn}, AVOD~\cite{ku2018avod}, and F-PointNet~\cite{qi2018frustum}. PointRCNN is LiDAR only detector while AVOD and F-PointNet required both LiDAR point clouds and camera images as input.
For all detectors, we directly use the same hyper-parameters that are released by the authors.

The 3D object detection performance is investigated on the most common vehicle, Car, in autonomous driving.
The evaluation metric is mean Average Precision (mAP) at $0.7$ IoU, the de-facto metric for object detection \cite{Geiger2012}.

\begin{table}[t]
  \centering
  \begin{tabular}{cclcc}
    \toprule
    &&& \multicolumn{2}{c}{Train} \\ \cmidrule{4-5}
    &&& KITTI & A*3D\\
    \midrule
    \parbox[t]{2mm}{\multirow{6}{*}{\rotatebox[origin=c]{90}{Validation}}}
    & \parbox[t]{2mm}{\multirow{3}{*}{\rotatebox[origin=c]{90}{KITTI}}}
    &  Easy     & \textbf{85.94} & 85.24 \\
    && Moderate & 76.02 & \textbf{80.11} \\
    && Hard     & 68.85 & \textbf{76.31} \\ \cmidrule{2-5}
    & \parbox[t]{2mm}{\multirow{3}{*}{\rotatebox[origin=c]{90}{A*3D}}}
    &  Easy     & 71.46 & \textbf{83.67} \\
    && Moderate & 61.76 & \textbf{76.02} \\
    && Hard     & 53.83 & \textbf{68.64} \\
    \bottomrule
  \end{tabular}
  \caption{
    Results of PointRCNN~\cite{shi2019pointrcnn} on 3D Car detection (mAP), with cross train-validation between KITTI and A*3D.
    Best results on each validation set are marked in bold.
  }
  \label{tab:density}
\vspace*{-5mm}
\end{table}

\subsection{Object-density: Cross-dataset Evaluation}
\label{sec:density}
To showcase the need for high density dataset, in this study, we consider the cases of cross train-validation between KITTI and A*3D, and vice versa.
For KITTI, we use the popular train/val split from MV3D~\cite{chen2017mv3d}, which contains $3,712$ training samples and $3,769$ validation samples.
For A*3D, we pick $5,208$ frames with high number of object instances ($\textgreater{10}$) to be our training set.

Evaluation results are reported in Table~\ref{tab:density}.
A pre-trained model of PointRCNN on KITTI suffers almost a $15\%$ drop in mAP on A*3D validation set.
On the other hand, when trained on our high-density subset, PointRCNN achieves much better performance on the KITTI validation set, especially on \emph{Moderate} and \emph{Hard} with almost $10\%$ improvements.

Two interesting findings are drawn from Table~\ref{tab:density}: (1) there is a bias in the current KITTI benchmark, where high density scenarios and highly-occluded objects are under-represented; and (2) adding high density samples into the training set helps to correct this bias and improve the robustness of the model on challenging cases.
This also shows that a good performance on KITTI does not reflect the performance in challenging environment settings, which motivated us to build the A*3D dataset that is much more diverse and with more complex scenes for autonomous driving.

\subsection{High object-density vs. Low object-density}
Results in Section~\ref{sec:density} show that adding high-density samples improves the performance and robustness of the 3D detector.
However, this leads to other questions such as how much additional training data is enough and what are the effects of adding more and more training data with different density settings.
Answering these questions may give valuable insight into what should be improved in either data configurations or algorithms for the current 3D object detection pipeline.

To answer these questions, we design an experiment where the PointRCNN network is trained with different training configurations extracted from A*3D: (A) training with high-density samples only, (B) training with low-density samples only, and (C) training with mix of high and low density samples.
For each configuration, we build different training data sets by gradually adding the number of samples.
More specifically, for the first and second configurations, we varies the training from $1$k to $5$k samples.
For the mix configuration, we keep the number of high-density samples at $1$k, and sequentially add more low-density samples into the set.

\begin{figure}[t]
  \centering
  \includegraphics[height=0.65\linewidth]{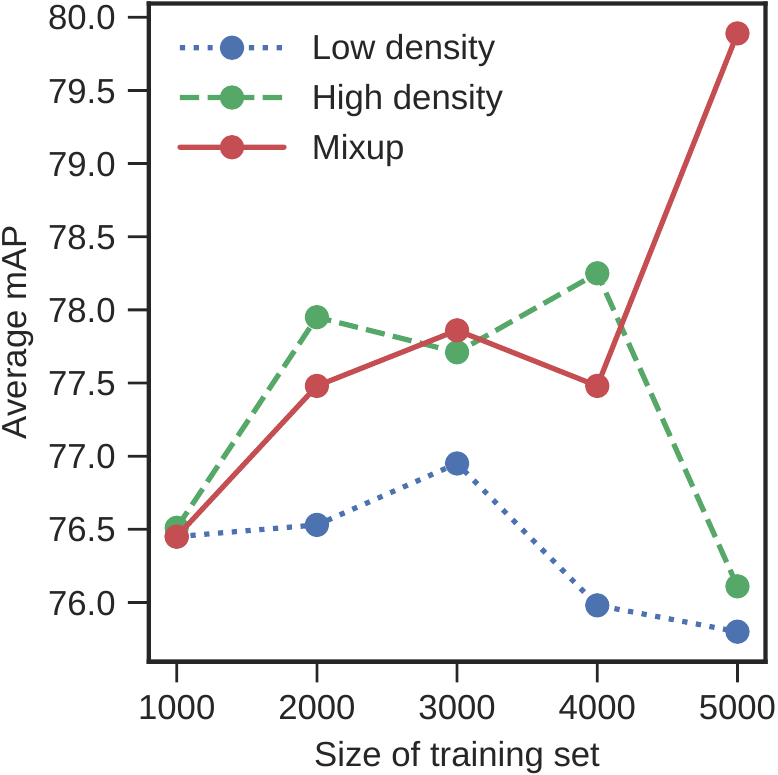}
  \caption{
     Results of PointRCNN on 3D Car detection (mAP over three  difficulty levels Easy/Moderate/Hard), with different training configurations on object density:
     training with \textbf{Low density} samples only, training with \textbf{High density} samples only, and training with the \textbf{Mix} of high and low object-density samples.
  }
  \label{fig:mAP}
\vspace*{-5mm}
\end{figure}

Fig.~\ref{fig:mAP} shows the results of PointRCNN on 3D Car detection (averaged mAP over three  difficulty levels Easy/Moderate/Hard), with different training configurations on object density.
Two interesting observations emerge from these results.
First, when increasing the training data, the performance improvements are marginal, suggesting that there is a law of diminishing returns at work.
This means that in order to boost the performance of a 3D object detector, we may need to put more efforts into algorithmic advances instead of increasing data quality and quantity.
Second, the best result comes from mixing high and low density samples.
This shows that training entirely on ``hard'' samples is not beneficial, as the easy samples probably can act as a form of regularization.
However, determining the ratio between easy and hard samples is still an open problem.

\subsection{Day-time vs. Night-time}
In this section, we investigate the effects of different lighting conditions (day-time vs. night-time) on the performance of 3D object detection systems with both LiDAR point clouds and camera images (from left camera) from A*3D.
Although several datasets have already provided night-time driving data~\cite{choi2018kaist,caesar2019nuscenes}, we are the first to provide a systematic study on the effects of night-time on 3D object detection systems.
Here we focus on methods that require or depend on RGB images for 3D detection, namely AVOD~\cite{ku2018avod} and F-PointNet~\cite{qi2018frustum}.
For F-PointNet, since the authors do not release their 2D detector model, we train a Mask R-CNN~\cite{he2017maskrcnn} model from scratch as our 2D detector.

We perform our training on the A*3D dataset, using the same train/validation split as detailed in Section~\ref{sec:density}, i.e., $5,208$ training and $1,500$ validation samples.
Both train and validation sets are further decomposed into day-time and night-time subsets respectively, based on their timestamps.
The day/night ratio is $2:1$ in both the train and validation sets.
To study the effects of different lighting conditions, we train three different models: one model using only images at day-time, one model using only images at night-time, and one model using both day and night images.

Table~\ref{tab:daynight} shows the evaluation results of AVOD and F-PointNet on A*3D with different lighting conditions.
We observe that (1) the models trained on both day and night data have consistent performance across different modalities and (2) specialized models can also achieve comparable performance in their own modality, especially for models trained with night data only which is with 3 times fewer training samples.
This leads to an interesting question that whether we should use a general model for detection or different specialized models depends on the weather and lighting conditions.

\begin{table}[t]
  \centering
  \setlength{\tabcolsep}{4pt}
  \begin{tabular}{cclcccccc}
    \toprule
    &&& \multicolumn{6}{c}{Train} \\\cmidrule{4-9}
    &&& \multicolumn{3}{c}{F-PointNet~\cite{qi2018frustum}} & \multicolumn{3}{c}{AVOD~\cite{ku2018avod}} \\ \cmidrule{4-9}
    &&& Day & Night & Both & Day & Night & Both \\
    \midrule
    \parbox[t]{2mm}{\multirow{9}{*}{\rotatebox[origin=c]{90}{Validation}}}
    & \parbox[t]{2mm}{\multirow{3}{*}{\rotatebox[origin=c]{90}{Day}}}
    &  Easy     & 81.91 & 74.24 & 80.40 & 85.87 & 77.63 & 86.52 \\
    && Moderate & 75.60 & 58.87 & 73.67 & 77.84 & 68.23 & 78.36 \\
    && Hard     & 68.40 & 50.27 & 67.17 & 76.71 & 66.80 & 70.48 \\ \cmidrule{2-9}
    & \parbox[t]{2mm}{\multirow{3}{*}{\rotatebox[origin=c]{90}{Night}}}
    &  Easy     & 75.82 & 81.81 & 82.15 & 85.40 & 85.03 & 85.19 \\
    && Moderate & 67.47 & 73.62 & 73.63 & 76.63 & 76.17 & 77.38 \\
    && Hard     & 59.34 & 67.49 & 67.24 & 69.77 & 69.61 & 76.91 \\ \cmidrule{2-9}
    & \parbox[t]{2mm}{\multirow{3}{*}{\rotatebox[origin=c]{90}{Both}}}
    &  Easy     & 81.88 & 74.98 & 80.60 & 85.68 & 77.75 & 86.15 \\
    && Moderate & 74.65 & 65.42 & 73.48 & 77.55 & 74.62 & 78.09 \\
    && Hard     & 67.72 & 57.76 & 67.23 & 76.61 & 67.83 & 77.23 \\
    \bottomrule
  \end{tabular}
  \caption{
    Results of F-PointNet and AVOD on 3D Car detection (mAP), with different train/validation configurations on lighting conditions.
  }
  \label{tab:daynight}
\end{table}

\begin{section}{Conclusion and Future Work}
We propose a novel multimodal A*3D dataset (RGB images and LiDAR data) for detection tasks for autonomous driving in challenging environments, addressing the gap in existing datasets in the literature.
A detailed benchmark on our dataset provides valuable insights about the limitations and robustness of the current state-of-the-art 3D object detection methods such as - training algorithms on high complexity and density data improves the robustness and accuracy of these algorithms; best performance is achieved by training on a combination of low and high density samples; and the performance of models in different conditions and times of the day is correlated with training in respective weather and lighting conditions. 
In future, we aim to answer the open questions identified in the paper by performing a more extensive evaluation of novel detection algorithms on A*3D and existing autonomous driving datasets.
We also intend to use this highly complex and dense A*3D dataset to develop more accurate algorithms for 3D object detection tasks related to autonomous driving in real-world settings.
\end{section}

\newpage
{
  \bibliographystyle{IEEEtran}
  \bibliography{ref}
}

\end{document}